\documentclass[letterpaper, 10 pt, conference]{ieeeconf}  

\IEEEoverridecommandlockouts                              
\usepackage{amsmath}
\usepackage{booktabs}
\usepackage[pdftex]{graphicx}
\usepackage{hyperref}
\hypersetup{colorlinks=true, citecolor=black, linkcolor=black}
\hypersetup{urlcolor=black}
\hypersetup{bookmarksnumbered=true}
\hypersetup{plainpages=false}
\usepackage[all]{hypcap} 

\usepackage{textcomp}

\title{\LARGE \bf
Stochastic Collection and Replenishment (SCAR): Objective Functions
}

\author{Andrew W Palmer, Andrew J Hill and Steven J Scheding$^{1}$
\thanks{This work has been supported by the Rio Tinto Centre for Mine Automation and the Australian Centre for Field Robotics, University of Sydney, Australia.}%
\thanks{$^{1}$The authors are with the Australian Centre for Field Robotics, University of Sydney, Australia. Email addresses: {\tt\small \{a.palmer;a.hill;s.scheding\}@acfr.usyd.edu.au}}%
\thanks{\textcopyright 2013 IEEE. Personal use of this material is permitted. Permission from IEEE must be obtained for all other uses, in any current or future media, including reprinting/republishing this material for advertising or promotional purposes, creating new collective works, for resale or redistribution to servers or lists, or reuse of any copyrighted component of this work in other works.}%
}

\begin{document}
\maketitle
\thispagestyle{empty}
\pagestyle{empty}

\begin{abstract}
This paper introduces two objective functions for computing the expected cost in the Stochastic Collection and Replenishment (SCAR) scenario. In the SCAR scenario, multiple user agents have a limited supply of a resource that they either use or collect, depending on the scenario. To enable persistent autonomy, dedicated replenishment agents travel to the user agents and replenish or collect their supply of the resource, thus allowing them to operate indefinitely in the field. Of the two objective functions, one uses a Monte Carlo method, while the other uses a significantly faster analytical method. Approximations to multiplication, division and inversion of Gaussian distributed variables are used to facilitate propagation of probability distributions in the analytical method when Gaussian distributed parameters are used. The analytical objective function is shown to have greater than 99\% comparison accuracy when compared with the Monte Carlo objective function while achieving speed gains of several orders of magnitude. 
\end{abstract}

\section{INTRODUCTION}
Resources such as fuel, battery charge and data storage space are a necessary component of autonomous systems. However, autonomous agents have a finite capacity for these resources and thus at some point will require the resource to be replenished, particularly when persistent autonomy is the goal. The aims of this paper are to introduce the Stochastic Collection and Replenishment (SCAR) scenario, and to present and compare two SCAR objective functions---a Monte Carlo method and a faster analytical method.

The SCAR scenario consists of two inverse problems---collection and replenishment. Collection refers to the case where user agents in the field are collecting a resource such as data or physical samples, whereas the replenishment case refers to agents using a resource such as fuel or battery charge. In the collection case, one or more collection agents travel to the field agents and retrieve the resource that the user agents have gathered. In the replenishment case, the replenishment agents travel to the user agents and replenish the resources of the user agents. In both cases, the aim of SCAR is to allow the user agents to continue operating without returning to base. 

Several replenishment scenarios exist in the literature---aerial refuelling of a fleet of planes or Unmanned Aerial Vehicles (UAVs) \cite{Jin2006a}, \cite{Jin2006b}, \cite{Kaplan2012}, \cite{Kaplan2013}, refuelling of a constellation of satellites \cite{Shen2002}, \cite{Tsiotras2005}, and recharging of UAVs using Unmanned Ground Vehicles (UGVs) \cite{Mathew2013}. The main collection scenario in the literature is the use of a data mule to collect data from a sensor field \cite{Tekdas2009}, \cite{Bhadauria2011}, \cite{Tekdas2012}. The above scenarios are formulated as an NP-hard optimisation problem, and the methods proposed for solving them include heuristics, simulated annealing, dynamic programming, and branch and bound. The objective functions generally revolve around minimising factors such as the total time or tardiness of the schedule, and the total fuel used or total distance travelled by the replenishment agent. 

The above literature have several limitations. Firstly, the replenishment agents are assumed to have sufficient capacity to fully fill all of the agents. However, when persistent autonomy is considered, it is clear that the replenishment agents themselves will have to be replenished at some point. Secondly, it is assumed that each user agent is replenished only once. Again, it is necessary to consider replenishing the user agents multiple times when performing persistent tasks. Finally, the agent parameters are treated as deterministic. In a practical system, many parameters such as speed and usage rates are stochastic in nature. Disregarding this uncertainty can lead to selection of suboptimal schedules from a risk point of view \cite{Kuindersma2012}. 

The SCAR scenario aims to address the above limitations by incorporating replenishment of the replenishment agent, multiple replenishments of the user agents, and uncertainty in the agent parameters. An open issue in optimisation of stochastic systems is estimation of the expected cost. A popular approach in the literature for calculating an expected cost has been scenario-based methods such as Monte Carlo simulation \cite{Bassett1997, Lin2004}. This is simple to implement, but can be computationally expensive. Gaussian quadrature integration was used in \cite{Ierapetritou1996} to generate an expected cost, while other authors have attempted to transform the stochastic problem into a deterministic problem using chance constraints \cite{Orcun1996} or conservative estimates of the uncertain parameters \cite{Bertsimas2003}. 


The main contribution of this paper is the formulation of two methods for calculating the expected cost in a SCAR scenario. The first is a Monte Carlo method that uses parameter values sampled from probability distributions to estimate an expected cost, while the second is an analytical solution that incorporates the entire probability distribution. The SCAR problem is firstly defined in Section \ref{s:probdef}, and mathematical models of the SCAR scenario are developed in Section \ref{s:modelform}. A deterministic model is developed for use in the Monte Carlo method, and a stochastic model for use in the analytical method. These methods are valid for any probability distribution, and are compared in this paper using Gaussian distributed variables. Section \ref{s:approxmeth} introduces a number of approximation methods for working with Gaussian distributed variables, including a new approximation to the Inverse Gaussian Distribution. A computational study of the objective functions is undertaken in Section \ref{s:compstudy}, followed by some concluding remarks in Section \ref{s:conc}. 


\section{PROBLEM DEFINITION} \label{s:probdef}
In the SCAR scenario there are multiple user agents, which have a limited capacity of a resource. In the replenishment problem, the user agents use the resource over time, while one or more replenishment agents travel around replenishing the supplies of the user agents. To replenish a user agent, the replenishment agent must travel to the user agent, setup before beginning replenishment, replenish the user agent, and then packup before it can travel to the next user agent. The user agent may continue to use the resource while being replenished. The inverse problem of the user agents collecting the resource is identical except for the direction of the resource flow. 

The replenishment agents also have a limited capacity of the resource and must periodically travel to a replenishment point to be replenished. The fleet of user agents is heterogeneous, as is the fleet of replenishment agents. The parameters of the agents such as usage rate, replenishment rate, speed, setup and packup time are stochastic. The user agents may be positioned apart from one another, and the motion of all agents may be constrained by roads or obstacles. 

Both user and replenishment agents can be replenished multiple times in a single schedule. Any user agent can be replenished by at most one replenishment agent at any time, and any replenishment agent can only service at most one user agent at any time. Preemption is not considered; i.e. once a replenishment agent has begun replenishing a user agent, it must continue until either the user agent is fully replenished or the replenishment agent has exhausted its supply of the resource. 

User agents operate in their own small operational areas such that they are unable to reach the replenishment point. As the distances travelled by the replenishment agent are significantly larger than the size of these operational areas, any variations in the travel times of the replenishment agent due to the movements of the user agents can be accounted for in the uncertainty of the setup and packup times of the replenishment agent. 

The consequence of a user agent running out of the resource may be either catastrophic (for example, the loss of the agent) which corresponds to a hard deadline, or non-catastrophic which corresponds to a soft deadline. A cost is incurred if the deadline is not met. A hard deadline incurs a step cost, while a soft deadline yields a linearly increasing cost. A single replenishment agent scenario with soft deadlines for the user agents has been assumed for this paper. 



\section{MODEL FORMULATION} \label{s:modelform}
This section presents a mathematical model for both the deterministic and stochastic single replenishment agent problems. The mathematics for the collection problem is identical if the resource considered is the free space remaining instead of the actual resource collected. 

\subsection{Parameters}
Each user agent, $i$, has the following parameters:
\begin{itemize}
\item $ u_{i,max} $ is the resource capacity
\item $ u_{i} $ is the current resource level
\item $ r_{i} $ is the resource usage rate
\end{itemize}
The replenishment agent, $a$, has the following parameters:
\begin{itemize}
\item $ u_{a,max} $ is the resource capacity
\item $ u_{a} $ is the current resource level
\item $ f_{a} $ is the resource replenishment rate into the user agent
\item $ t_{s,a} $ is the setup time
\item $ t_{p,a} $ is the packup time
\item $ v_{a} $ is the velocity
\end{itemize}
The replenishment point, $r$, has the following parameters:
\begin{itemize}
\item $ t_{s,r} $ is the setup time
\item $ t_{p,r} $ is the packup time
\item $ f_{r} $ is the resource replenishment rate into the replenishment agent
\end{itemize}

The distance between the replenishment agent and a user agent is denoted $ d_{a,i} $ and the distance between the replenishment agent and the replenishment point is denoted $ d_{a,r} $. In the following equations, an upper-case variable denotes a probability distribution. Units do not matter as long as they are consistent. 

\subsection{Deterministic Equations} \label{s:deterministic}
The cost, $ c $, of a schedule when using a soft deadline is a ratio between 0 and 1 which indicates what proportion of time the agents are empty. A value of 0 means that none of the user agents were empty at any point during the schedule, while a value of 1 means that all of the user agents were empty for the entire time required to complete the schedule. A schedule denotes a series of tasks for the replenishment agents in the order to be executed. There are two possible task types---replenishing a user agent, or replenishing the replenishment agent. The first involves the replenishment agent travelling to the user agent, while the second requires the replenishment agent to travel to the replenishment point. An example schedule for a single-replenishment agent, four-user agent system might be [0, 2, `r', 3, 1]. In this schedule, the replenishment agent would replenish user agents 0 and 2 before visiting the replenishment site, and would then replenish user agents 3 and 1. 

Each time a user agent is replenished, the time that it was empty up until that point must be calculated. For each task replenishing a user agent, the time empty in a segment, $t_{segment,i}$, can be calculated as:
\[t_{segment,i} = max(0, t_{b,i} - t_{e,i})\]
where

\begin{itemize}
\item $t_{b,i}$ is the time at which the replenishment agent began replenishing user agent, $i$
\item $t_{e,i}$ is the time when user agent, $i$, will fully deplete its resources
\end{itemize}

Once the schedule has been executed, the time empty for each user agent must be calculated based on when it was last replenished. So for each user agent, the time empty at the end of the schedule, $t_{end,i}$ is:
\[t_{end,i} = \max (0, t_{max} - t_{e,i})\]
where $t_{max}$ is the time that the schedule will finish being executed. For a soft deadline, the cost of the schedule is then defined as:
\begin{equation} \label{eq:costcalcd}
c = \frac{\displaystyle\sum_{i=0}^{n} t_{segment,i} + \sum_{i=0}^{n} t_{end,i}}{n t_{max}}
\end{equation}
where $n$ is the total number of user agents. The $n t_{max}$ term normalises the cost such that $ 0 \leq c \leq 1 $. A hard deadline would be represented by a cost proportional to the number of user agents that ran out of the resource. For each user agent, the time till empty can be calculated as:
\begin{equation} \label{eq:emptytimed}
t_{e,i} = \frac{u_{i}}{r_{i}}
\end{equation}

In most cases, $u_{i} = u_{i,max}$. However, if the replenishment agent did not have sufficient resources to completely replenish the user agent previously, then $u_{i}$ will be lower than $u_{i,max}$. 

The time until a user agent is replenished depends on when the replenishment agent last finished servicing a user agent. Let $ t_{f,i} $ be the time that the replenishment agent has finished replenishing user agent $i$, has packed up and is ready to travel to the location of the next task. Then the time that user agent $i+1$ is replenished, $t_{b,i+1}$, will depend on $ t_{f,i}$:
\begin{equation} \label{eq:timefilld}
t_{b,i+1} = t_{f,i} + \frac{d_{a,i}}{v_{a}} + t_{s,a}
\end{equation}

Then $t_{f,i+1}$ can be calculated as:
\begin{equation} \label{eq:timeemptyd}
t_{f,i+1} = t_{b,i+1} + t_{r,i+1} + t_{p,a}
\end{equation}

$t_{r,i}$ is the time that it takes to replenish user agent $i$ and depends on the current resource level, the resource usage rate and the replenishment rate:
\[t_{r,i} = \frac{u_{i,max} - u_{i}}{f_{a}} + \frac{t_{r,i} r_{i}}{f_{a}} = \frac{u_{i,max} - u_{i}}{f_{a} - r_{i}}\]

However, if the replenishment agent does not have sufficient resources to fully replenish the user agent, then the time to replenish the user agent will be dependent on the resource level of the replenishment agent:
\begin{equation} \label{eq:filld}
t_{r,i} = \min(\frac{u_{a}}{f_{a}},\frac{u_{i,max} - u_{i}}{f_{a} - r_{i}})
\end{equation}

The resource level in user agent, $i$, after being replenished, $u_{i}^{t_{1}}$, will be:
\[u_{i}^{t_{1}} = \min (u_{i,max}, u_{i}^{t_{0}} + t_{r,i}(f_{a} - r_{i}))\]
where $u_{i}^{t_{0}}$ is the resource level before being replenished. The resource level in the replenishment agent will then be:
\[u_{a}^{t_{1}} = \max (0, u_{a}^{t_{0}} - t_{r,i} f_{a})\]

For an `r' task, the total time $t_{f,a}$ will include the travel time to the replenishment site, setup and packup times of the replenishment site, and the time to replenish the replenishment agent:
\begin{equation} \label{eq:fill_r_agent_d}
t_{f,a} = t_{f,i} + \frac{d_{a,r}}{v_{a}} + t_{s,r} + \frac{u_{a,max} - u_{a}}{f_{r}} + t_{p,r}
\end{equation}

\subsection{Stochastic Equations} \label{s:stohasticeq}
As noted previously, a number of the parameters of the system are stochastic in practice, and a deterministic cost calculation does not capture the inherent risk of a particular schedule. Using just the expected values of the parameters, the cost for two different schedules may be zero. Once uncertainty is taken into account, one schedule could be found to be more risky than the other.

The uncertainty can be incorporated by calculating an expected cost using the probability distributions of the parameters. The most straight-forward method to do this is a Monte Carlo cost analysis where the deterministic cost calculation is run multiple times using values sampled from the probability distributions. The quality of the cost estimate will improve on average as more samples are used. However, using a large number of samples has a significant impact on computation time when running a schedule optimisation. 

A faster method of calculating the expected cost is to explicitly calculate the probability distribution for the time empty at each step of the schedule and calculate the expected cost from these probability distributions. The time that the user agent is empty is given by:
\[T_{segment,i} = T_{b,i} - T_{e,i}\]

As the time empty can only be  $\geq 0$, the expected value of the time empty is given by:

\begin{equation} \label{eq:integralcost}
E[T_{segment,i}] = \int\limits_{0}^{\infty } \! x p(T_{segment,i}) \ \textrm{d}x
\end{equation}
where $p(T_{segment,i})$ denotes the probability density function of $T_{segment,i}$. Similarly at the end of the schedule:
\[T_{end,i} = T_{max} - T_{e,i}\]
\[E[T_{end,i}] = \int\limits_{0}^{\infty } \! x p(T_{end,i}) \ \textrm{d}x\]

Therefore the expected cost of the schedule for a soft deadline is:
\begin{equation} \label{eq:expectedcost}
E[C] = \frac{\displaystyle\sum_{i=0}^{n} E[T_{segment,i}] + \sum_{i=0}^{n} E[T_{end,i}]}{n E[T_{max}]}
\end{equation}
where $E[T_{max}]$ is given by the mean value of the distribution $p(T_{max})$. For a hard deadline, the cost would reflect the probability of the user agents running empty rather than the total time that they are empty. 

Most of the following equations are similar to the deterministic equations from Section \ref{s:deterministic}, using probability distributions rather than explicit values. Equation (\ref{eq:emptytimed}) becomes:
\[T_{e,i} = \frac{U_{i}}{R_{i}}\]

Equation (\ref{eq:timefilld}) becomes:
\[T_{b,i+1} = T_{f,i} + \frac{d_{a,i}}{V_{a}} + T_{s,a}\]

Let $T_{m,i+1}$ be the time that the user agent is finished being replenished:
\begin{equation}\label{eq:timefinishedfilling}
T_{m,i+1} = T_{b,i+1} + T_{r,i+1}
\end{equation}

Using (\ref{eq:timefinishedfilling}), (\ref{eq:timeemptyd}) becomes:
\begin{equation}
T_{f,i+1} = T_{m,i+1} + T_{p,a}
\end{equation}

Equation (\ref{eq:filld}) cannot be simply modified like the previous equations as taking the minimum between two probability distributions would lead to a discontinuous probability distribution. Therefore, a series of steps are required to calculate the time that it takes to replenish the user agent. Firstly, calculate the resource level of the user agent at time $T^{t_{1}}_{b,i}$:
\[U_{i}^{t_{1}} = U_{i}^{t_{0}} - R_{i} (T^{t_{1}}_{b,i} - T^{t_{0}}_{m,i})\]
where $T^{t_{0}}_{m,i}$ is the time that the previous replenishment of the user agent was finished at. Depending on the choice of probability distribution, some of the distribution will extend into the negative part of the domain. However, a negative level of resource does not make sense, so steps should be taken to minimise the probability of a negative value. Similarly, a resource level greater than the maximum resource level is impossible and should be accounted for. Section \ref{s:adjusting} discusses a possible adjustment technique. A \# will be used to denote variables that have been adjusted to match the physical constraints. 

The next step is to calculate the time it will take for the replenishment agent to replenish the user agent assuming that the replenishment agent has sufficient resource levels to achieve this, $T_{r,i}$:
\[T_{r,i} = \frac{u_{i,max} - U_{i}^{t_{1}\#}}{F_{a} - R_{i}}\]

Taking into account the quantity of resource used by the user agent during the replenishment, the amount of resource to be replenished, $Q_{i}$, is given by:
\[Q_{i} = u_{i,max} - U_{i}^{t_{1}\#} + T_{r,i} R_{i}\]

However, the replenishment agent may not have sufficient resources to fully replenish the user agent, so $Q_{i}$ must be adjusted against $U_{a}$. The actual replenishment time can then be calculated as:
\[T_{r,i} = \frac{Q_{i}^{\#}}{F_{a}}\]

The resource level of the user agent after being replenished, $U_{i}^{t_{2}}$, is given after adjustment against $u_{i,max}$ by:
\[U_{i}^{t_{2}} = (U_{i}^{t_{1}\#} + U_{a} - T_{r,i} R_{i})^{\#}\]

$U_{a}$ is used instead of $T_{r,i}F_{a}$ to ensure that the adjusted distribution has an appropriate shape. The resource level of the replenishment agent is then given by:
\[U_{a}^{t_{2}} = (U_{a}^{t_{1}} - Q_{i}^{\#})^{\#}\]

Finally, (\ref{eq:fill_r_agent_d}) becomes:
\begin{equation} \label{eq:fill_r_agent_s}
T_{f,a} = T_{f,i} + \frac{d_{a,r}}{V_{a}} + T_{s,r} + \frac{u_{a,max} - U_{a}}{F_{r}} + T_{p,r}
\end{equation}

\section{APPROXIMATION METHODS} \label{s:approxmeth}
It is assumed that the stochastic parameters can be described by independent probability distributions. While any probability distribution is valid, in this paper the parameters have been modelled with Gaussian distributions as these are relatively easy to work with. Additions and subtractions of Gaussian distributed variables result in a Gaussian distribution, whereas other operations such as division, multiplication and inversion result in non-Gaussian distributions. To facilitate the propagation of the distributions throughout the objective function, it is convenient to approximate the non-Gaussian distributions as Gaussian distributions. The following subsections discuss these approximation methods. It should be noted that the probability distributions for quantities such as the time that the user agent was last replenished at and the time taken to replenish the user agent are not independent and there will be some covariance between them. However, for this paper the covariance between all calculated distributions has been assumed to be zero. It will be demonstrated later in this paper that these are acceptable approximations and assumptions. 

\subsection{Inverse Gaussian Distribution} \label{s:inverse}
An approximation of the probability distribution resulting from the inverse of a Gaussian distributed variable was not found in the examined literature. Therefore, an approximation was formulated through a numerical study. Consider:
\[
I = \frac{c}{G}
\]
where $I$ is an inverse Gaussian distributed variable, $G$ is a Gaussian distributed variable described by mean, $\mu_{G}$, and standard deviation, $\sigma_{G}$, and $c$ is a constant. The distribution of $I$ is then approximately described by a Gaussian distribution with mean, $\mu_{I}$, and standard deviation, $\sigma_{I}$, where:
\begin{equation}
\mu_{I} = \frac{C}{\mu_{G}} \qquad \sigma_{I} = \frac{\sigma_{G} C}{\mu_{G}^2}
\end{equation}

\begin{figure}
  \centering
    \includegraphics[width=0.48\textwidth]{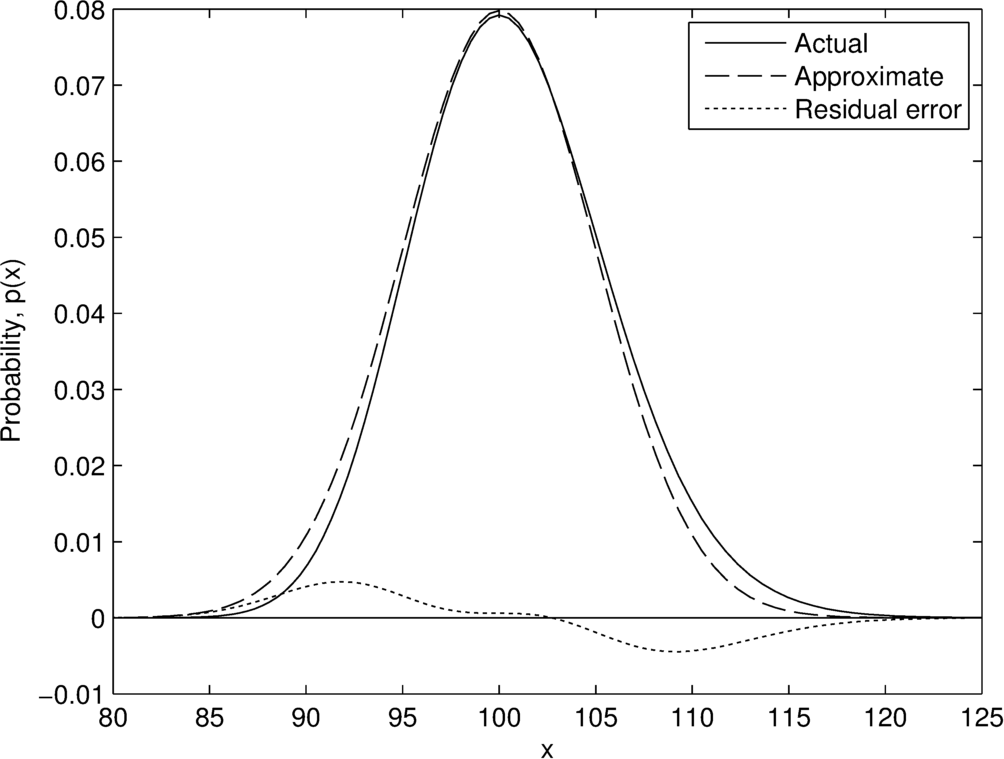}
  \caption{Inverse Gaussian Distribution and Approximation for $\mu_{G} / \sigma_{G} = 20$}
  \label{f:inversegauss}
\end{figure}

\begin{figure}
  \centering
    \includegraphics[width=0.48\textwidth]{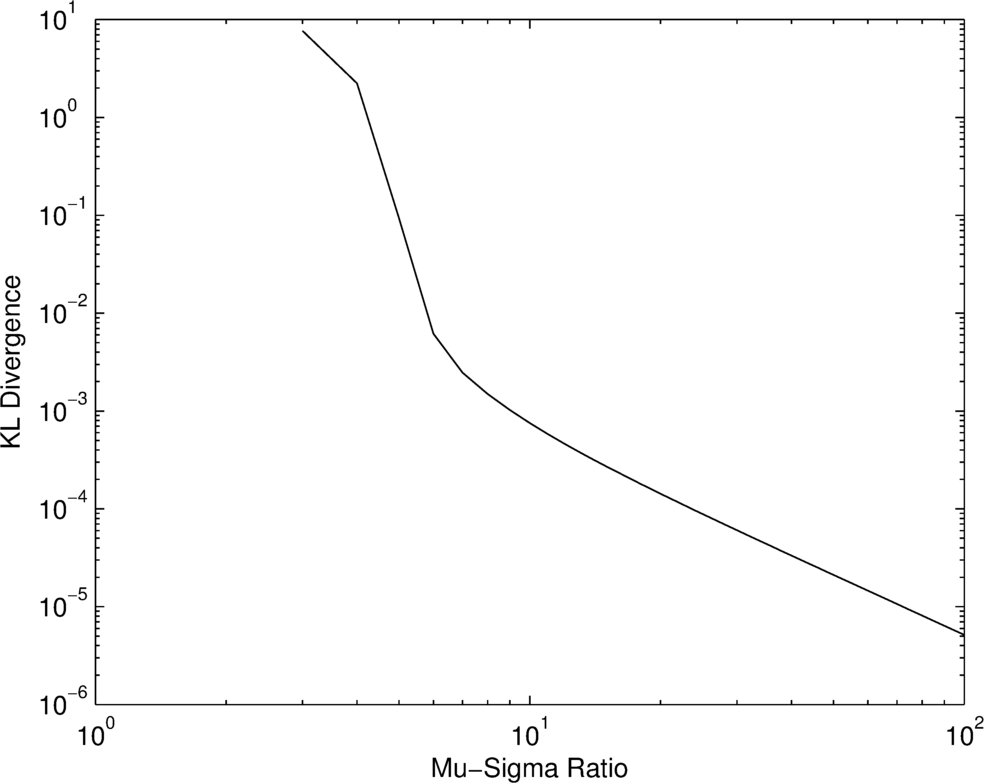}
  \caption{KL Divergence for the Inverse Gaussian Approximation}
  \label{f:kldivergence}
\end{figure}

Fig. \ref{f:inversegauss} shows the inverse Gaussian distribution, the resultant approximation and the residual error between the two distributions for an initial $\mu_{G} / \sigma_{G}$ ratio of 20. The Kullback-Leibler (KL) Divergence \cite{Kullback1951} for various ratios is shown in Fig. \ref{f:kldivergence}. The KL Divergence is a measure of the difference between two distributions, and as it nears 0, the approximation approaches the actual distribution. Fig. \ref{f:kldivergence} shows that the approximation improves as the $\mu_{G} / \sigma_{G}$ ratio increases.

\subsection{Gaussian Ratio Distribution}
A method for approximating a ratio of Gaussian distributed variables is given in \cite{Marsaglia2006}. A ratio distribution, $R$:
\[
R = \frac{E}{F}
\]
with $\mu_{E}$, $\sigma_{E}$, $\mu_{F}$, $\sigma_{F}$ and $\rho = 0$ can be approximated with a Gaussian distribution where:
\[
r = \frac{\sigma_{F}}{\sigma_{E}} \textrm{,} \ a =  \frac{\mu_{E}}{\sigma_{E}} \ \textrm{and} \ b =  \frac{\mu_{F}}{\sigma_{F}}
\]
\begin{equation} \label{eq:ratiomu}
\mu_{R} = \frac{a}{r(1.01b - 0.2713)}
\end{equation}
\begin{equation} \label{eq:ratiosigma}
\sigma_{R} = \frac{1}{r}\sqrt{\frac{a^{2}+1}{b^{2} + 0.108b - 3.795} - r^{2}\mu_{R}^{2}}
\end{equation}

This approximation is only valid for $a < 2.5$, $b > 4$ \cite{Marsaglia2006}. As $a \rightarrow \infty$, the ratio distribution approaches the inverse Gaussian distribution and can be approximated as per section \ref{s:inverse}. For situations where $a \ge 2.5$, the inverse Gaussian approximation method is used with $E$ treated as a constant, $e = \mu_{E}$. 

\subsection{Gaussian Product Distribution}
An approximation to the product of two Gaussian distributed variables is presented in \cite{Ware2003}. For a Gaussian product distribution, $M$:
\[
M = EF
\]
Then:
\begin{equation}
\mu_{M} = \mu_{E}\mu_{F}
\end{equation}
\begin{equation}
\sigma_{M}^{2} = \sigma_{E}^{2}\sigma_{F}^{2}(1 + \delta_{E}^{2} + \delta_{F}^{2})
\end{equation}
where
\[ \delta_{x} = \frac{\mu_{x}}{\sigma_{x}}\]

The authors note that the approximation improves as $\delta_{E}$ and $\delta_{F}$ become large. 

\subsection{Expected Value}
Equation (\ref{eq:integralcost}) gives the expected time empty for part of a schedule. Assuming the expected time empty is a Gaussian distribution with $\mu$ and $\sigma$, this becomes:
\[E[G] = \int\limits_{0}^{\infty } \! \frac{x}{\sigma\sqrt{2\pi}}e^{\frac{-(x-\mu)^{2}}{2\sigma^2}} \ \textrm{d}x\]

Solving the definite integral gives:
\begin{equation} \label{eq:expectedvalue}
E[G] = \frac{\mu}{2}(1+\textrm{erf}(\frac{\mu}{\sigma\sqrt{2}})) + \frac{\sigma}{\sqrt{2\pi}}e^{-\frac{\mu^{2}}{2\sigma^{2}}}
\end{equation}
where erf is the Gauss error function. 

\subsection{Adjusting Distributions} \label{s:adjusting}
Gaussian distributions have a domain of $(-\infty,\infty)$, and therefore there is some probability that a sampled value will be outside the physical limits. When considering quantities such as resource level for example, a negative value is not physically possible. Similarly, the maximum value is constrained by the capacity of the agent. A simple method was devised that shifts the mean and adjusts the standard deviation such that the majority of the distribution is within the logical bounds. If the distribution is within 3 standard deviations of 0, then:

\begin{itemize}
\item $\mu^{\#}$ is calculated using (\ref{eq:expectedvalue})
\item $\sigma^{\#} = \mu^{\#} / 3$
\end{itemize}

\begin{figure}
  \centering
    \includegraphics[width=0.48\textwidth]{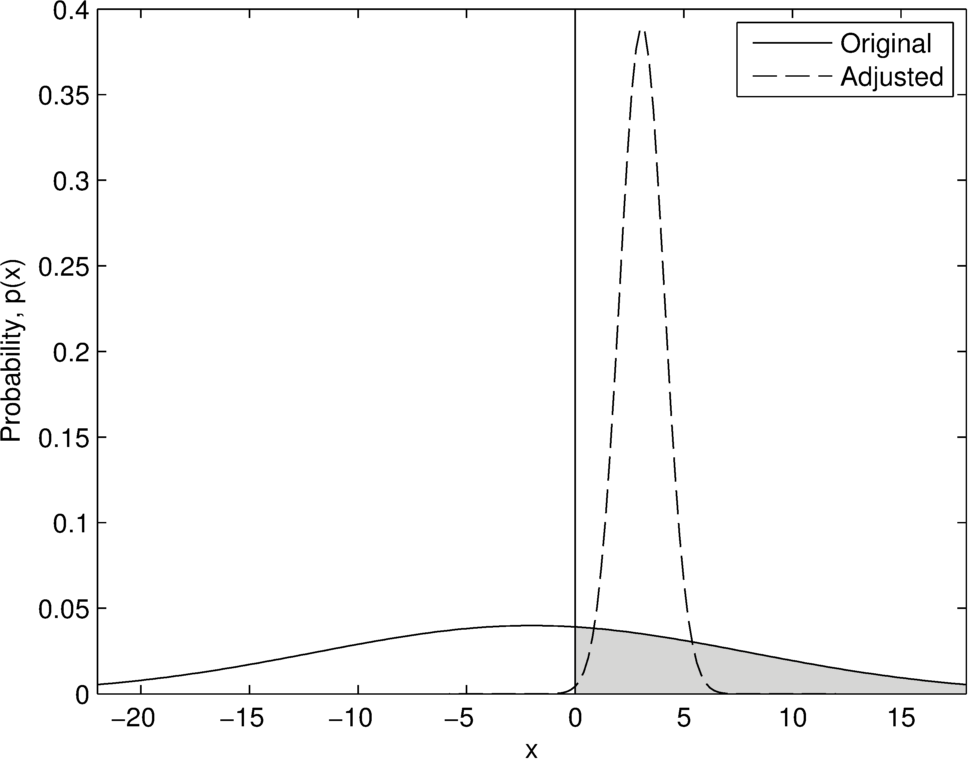}
  \caption{Original and Adjusted Distributions}
  \label{f:adjust}
\end{figure}

By keeping the mean 3 standard deviations away from 0, the probability of a negative value is 0.13\%. As an example, consider a future resource level which has been calculated as $\mu = -2$ and $\sigma = 10$. As shown by the shaded section in Fig. \ref{f:adjust}, 42.07\% of the probability currently lies in the positive domain. The expected positive value of this distribution is calculated using (\ref{eq:expectedvalue}) as 3.069. Therefore, the adjusted parameters are:

\[ \mu^{\#} = 3.069 \ \textrm{and} \ \sigma^{\#} = \mu^{\#} / 3 = 1.023 \] 

The adjusted distribution is shown in Fig. \ref{f:adjust}. As the initial distribution moves further into the negative part of the domain, the adjusted values will both approach 0. The following steps should be used to adjust against the maximum level of the resource:

\begin{itemize}
\item Subtract the resource level from the maximum resource level
\item Perform the adjustment on the resultant distribution
\item Subtract the adjusted distribution from the maximum resource level
\end{itemize}

Note that this is just one possible method for adjusting the distribution. Using other distributions such as the Gamma distribution which does not go below 0 may provide a better approximation. However, the difficulty of working with both Gaussian and non-Gaussian distributions then arises. 

\section{COMPUTATIONAL STUDY} \label{s:compstudy}
\begin{figure}
  \centering
    \includegraphics[width=0.48\textwidth]{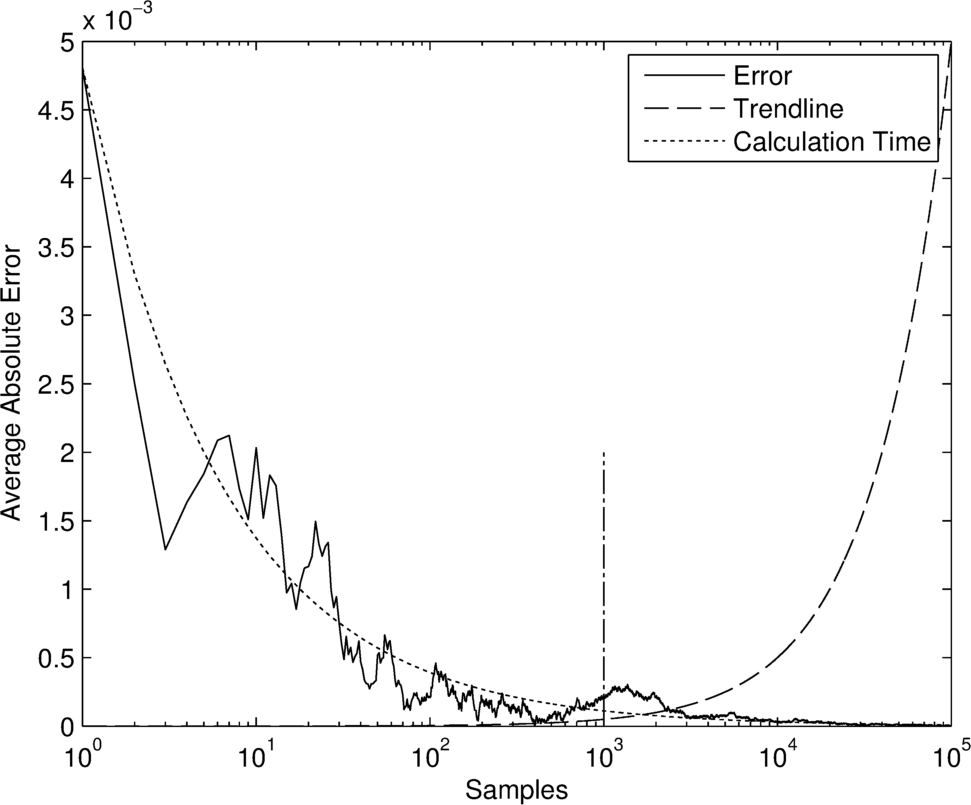}
  \caption{Average absolute error versus Monte Carlo samples. Calculation time shown as a guide. 1000 sample point marked. }
  \label{f:monte}
\end{figure}

\begin{figure}
  \centering
    \includegraphics[width=0.4\textwidth]{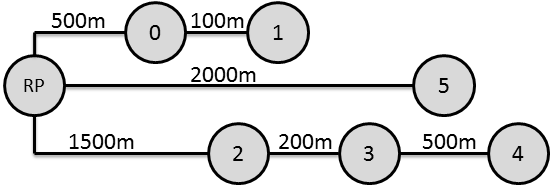}
  \caption{Layout showing the location of the Replenishment Point (RP) and User Agents}
  \label{f:layout}
\end{figure}

To test the relative performance of the objective functions, a Monte Carlo analysis was performed. For each randomly generated schedule, the expected cost was calculated using both a Monte Carlo approach with the equations detailed in Section \ref{s:deterministic}, and an analytical approach using the equations detailed in Section \ref{s:stohasticeq} combined with the approximation methods detailed in Section \ref{s:approxmeth}. Each Monte Carlo cost was calculated using 1000 samples from each of the parameter's probability distributions as a compromise between accuracy and computation time. This trade-off is shown in Fig. \ref{f:monte}. The scenario consisted of a single replenishment agent, a homogeneous set of 6 user agents with soft deadlines, and a single replenishment point. The distances between the user agents and the replenishment point are shown in Fig. \ref{f:layout}. The replenishment agent starts at the replenishment point. The parameters of the agents and the replenishment point are detailed in Tables \ref{t:param_r}, \ref{t:param_u} and \ref{t:param_p}. 

\begin{table}[b]
\caption{Replenishment Agent Parameters}
\label{t:param_r}
\centering
\begin{tabular}{ c  c  c }
\toprule
Parameter & Mean & Standard Deviation \\
\midrule
$u_{max}$ (units) & 5000 & 0 \\
$F$ (units/s) & 10 & 0.5 \\
$T_{s}$ (s) & 60 & 20 \\
$T_{p}$ (s) & 20 & 5 \\
$V$ (m/s) & 15 & 0.5 \\
\bottomrule
\end{tabular}
\end{table}

\begin{table}
\caption{User Agent Parameters}
\label{t:param_u}
\centering
\begin{tabular}{ c  c  c }
\toprule
  Parameter & Mean & Standard Deviation \\
  \midrule
  $u_{max}$ (units) & 1000 & 0 \\
  $R$ (units/s) & 0.5 & 0.05 \\
  \bottomrule
\end{tabular}
\end{table}

\begin{table}
\caption{Replenishment Point Parameters}
\label{t:param_p}
\centering
\begin{tabular}{ c  c  c }
\toprule
  Parameter & Mean & Standard Deviation \\
  \midrule
  $T_{s}$ (s) & 30 & 10 \\ 
  $T_{p}$ (s) & 10 & 1 \\ 
  $F$ (units/s) & 20 & 1 \\ 
  \bottomrule
\end{tabular}
\end{table}

\begin{figure}
  \centering
    \includegraphics[width=0.48\textwidth]{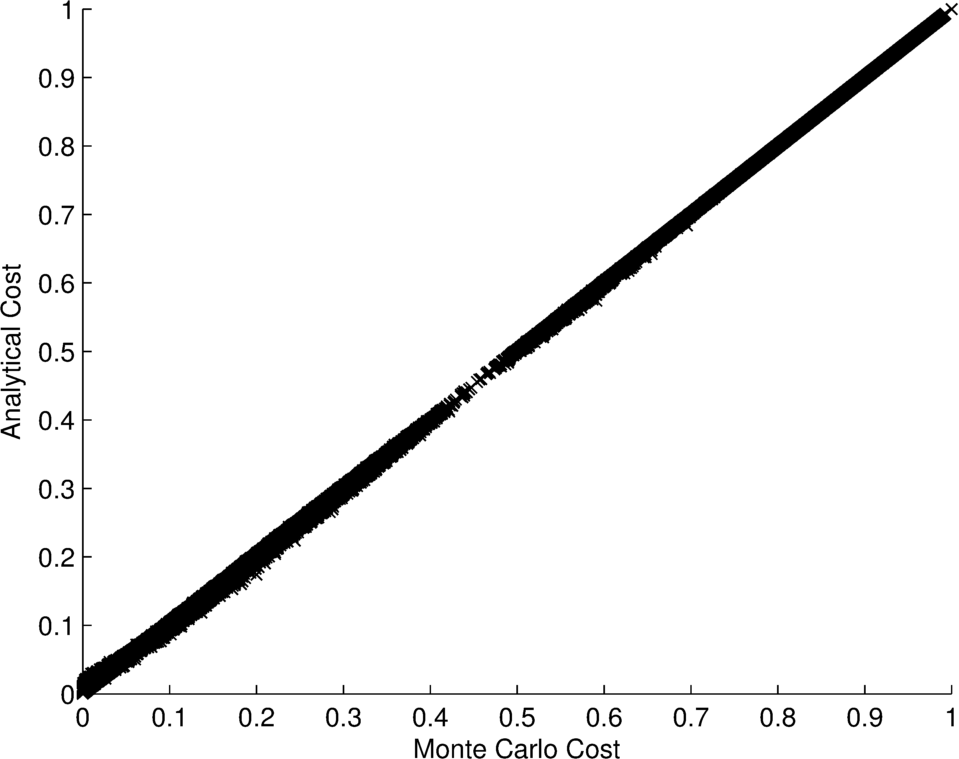}
  \caption{Analytical Cost versus Monte Carlo Cost}
  \label{f:results}
\end{figure}

A random selection of 90,000 schedules of between 5 and 10 tasks was generated with the only limitation being that consecutive tasks could not be the same. A schedule of [0,1,0,1,2] would be valid, while a schedule of [0,0,1,1,2] would not. The sample of 90,000 schedules represents around 0.1\% of the possible schedules, and this number was chosen to limit the run time of the Monte Carlo analysis. One third of the schedules were run with all agents starting at empty, one third with all agents starting at half capacity, and the final third with all agents starting at full capacity. Fig. \ref{f:results} shows that the analytical objective function provides a close approximation to the Monte Carlo generated cost. A linear trend line was fitted to the data with an equation of $y = 0.9983x + 0.0013$ and an $\textrm{R}^{2}$ value of 0.99996 (an $\textrm{R}^{2}$ value close to 1 indicates a good fit).

\begin{figure}
  \centering
    \includegraphics[width=0.48\textwidth]{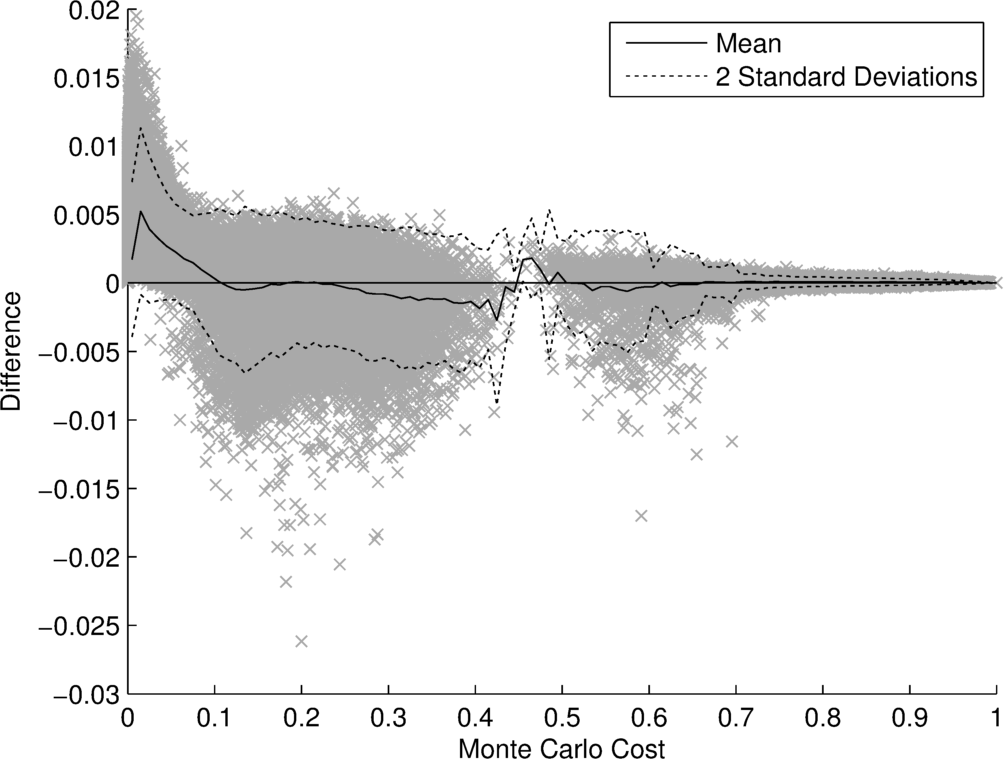}
  \caption{Analytical Cost minus Monte Carlo Cost versus Monte Carlo Cost for 6 User Agents}
  \label{f:error}
\end{figure}

For each of the initial conditions, the mean and standard deviation of the analytical costs minus the Monte Carlo costs is given in Table \ref{t:results1}. Fig. \ref{f:error} shows the differences plotted against the Monte Carlo cost. The error generally decreases as the cost approaches 1. Some of this error can be attributed to the use of only 1000 iterations for the Monte Carlo cost function. As shown in Fig. \ref{f:monte}, this error is on average around $0.5 \times 10^{-3}$. The rest of the error is likely due to the approximations and assumptions from Section \ref{s:approxmeth} that were used to calculate the analytical cost. For example, the assumption that the probability distributions are independent is not necessarily valid. However, based on these results, the effects are arguably negligible. 

\begin{table} [b]
\caption{Analytical Cost minus Monte Carlo Cost for 6 User Agents}
\label{t:results1}
\centering
\begin{tabular}{ c  c  c  }
\toprule
  Initial  & Mean & Standard   \\
  Conditions & ($\times 10^{-3}$) & Deviation ($\times 10^{-3}$)  \\
  \midrule
  Empty & 0.02 & 0.56 \\ 
  Half full & -0.10 & 2.79 \\ 
  Full & 2.02 & 2.72 \\ 
  \bottomrule
\end{tabular}
\end{table}

While these measures show the consistency between the two methods, the primary role of the objective function is to discriminate between good and bad schedules. To test the accuracy of the analytical objective function, over 5 billion comparisons were made between schedules. For two schedules A and B, a correct result is recorded if both the Monte Carlo method and analytical method agree on which of A and B is better. Comparisons where the Monte Carlo cost for A and B were both equal to 0 were not included as the analytical method will return a very small but non-zero number for these schedules. The comparisons were run on the results from each set of initial conditions individually as well as on the entire set of results. The results of these comparisons are given in Table \ref{t:results2}. 

\begin{table} [b]
\caption{Comparison Accuracy for 6 User Agents}
\label{t:results2}
\centering
\begin{tabular}{ c  c  c }
\toprule
  Initial  & Number of &  Correct  \\
  Conditions & Comparisons &   \\
  \midrule
  Empty & 446129609 & 99.97\%  \\ 
  Half full & 449984570 &  99.05\%  \\ 
  Full & 342711946 & 97.65\%  \\ 
  All & 3938386683 & 99.66\% \\
  \bottomrule
\end{tabular}
\end{table}

The weighted average accuracy using the individual results is 99.00\%. The accuracy when the initial conditions are full is lower than for the other initial conditions. Table \ref{t:results3} shows that the mean difference in cost between schedules for the incorrect comparisons is very small in comparison to the overall mean difference for the full initial conditions. While it isn't 100\% accurate, for the cases in which it produces incorrect results the cost differences are possibly small enough that the schedules could be considered effectively identical. 

\begin{table}
\caption{Mean Difference Between Solutions for Full Initial Conditions}
\label{t:results3}
\centering
\begin{tabular}{ c  c  c}
\toprule
    & Monte Carlo  & Analytical \\
    & ($\times 10^{-3}$) & ($\times 10^{-3}$) \\
  \midrule
  All comparisons & 27.51  & 29.60 \\ 
  Incorrect comparisons & 1.02 & 1.80  \\ 
  \bottomrule
\end{tabular}
\end{table}

If all of the initial conditions are considered together, the accuracy is 99.66\%. This is a slightly unfair comparison as the initial conditions significantly influence the cost. However, the result highlights the discrimination accuracy of the method, in particular for cases of more extreme cost variation. 

\addtolength{\textheight}{-1.5cm}

A similar study was performed on a larger scenario involving a heterogeneous set of 20 user agents and schedule lengths of between 16 and 20 tasks. Table \ref{t:results4} shows the error between the analytical cost and Monte Carlo cost. With the increased number of user agents and tasks, the mean error has increased slightly while the standard deviation has overall decreased. Table \ref{t:results5} shows that the comparison accuracy for the empty, half full and full cases has decreased by at most 0.81\%. The weighted average accuracy is 98.8\%, while the overall accuracy has not changed from 99.66\%. 

\begin{table} [b]
\caption{Analytical Cost minus Monte Carlo Cost for 20 User Agents}
\label{t:results4}
\centering
\begin{tabular}{ c  c  c  }
\toprule
  Initial  & Mean & Standard   \\
  Conditions & ($\times 10^{-3}$) & Deviation ($\times 10^{-3}$)  \\
  \midrule
  Empty & -0.40 & 1.31 \\ 
  Half full & -5.62 & 2.78 \\ 
  Full & 0.40 & 1.31 \\ 
  \bottomrule
\end{tabular}
\end{table}

\begin{table}
\caption{Comparison Accuracy for 20 User Agents}
\label{t:results5}
\centering
\begin{tabular}{ c  c  c }
\toprule
  Initial  & Number of &  Correct  \\
  Conditions & Comparisons &   \\
  \midrule
  Empty & 12497500 & 99.90\%  \\ 
  Half full & 12497500 & 98.65\%  \\ 
  Full & 5351411 & 96.84\%  \\ 
  All & 105342630 & 99.66\% \\
  \bottomrule
\end{tabular}
\end{table}

The main advantage of the analytical objective function is the computation time. Compared to the Monte Carlo method, the analytical method can be several orders of magnitude faster. Using a schedule with 10 tasks as an example, the Monte Carlo method using 1000 samples took 665ms to calculate the cost, while the analytical method took only 2ms. These times were calculated using the same hardware and both methods were implemented in Python by the same programmer. 

\section{CONCLUSIONS} \label{s:conc}

This paper introduced a framework for the SCAR scenario. Methods for approximating Gaussian probability distributions were discussed and a new method of approximating the inverse Gaussian distribution was presented. These were used to quickly calculate an analytical expected cost of a schedule. A computational study between the analytical objective function and a Monte Carlo generated expected cost showed that the analytical objective function was over 99\% accurate while computing several orders of magnitude faster in comparison to the Monte Carlo method. 

In future research, the new objective function will be used in conjunction with combinatorial optimisation techniques such as simulated annealing and branch and bound to improve the quality of results and speed of computation. Future work could include accounting for the covariance between distributions and performing a sensitivity analysis on the analytical method. Possible extensions to this work include using and collecting multiple resources such as fuel and data at the same time, moving the operational areas of the user agents at uncertain rates, and considering multiple replenishment agents. 



\bibliographystyle{IEEEtran}
\bibliography{IEEEabrv,ref}

\end{document}